\documentclass[10pt,twocolumn,letterpaper]{article}

\usepackage[pagenumbers]{cvpr} 

\usepackage{amsmath,amsfonts}
\usepackage{bm}
\usepackage{multirow}
\newcommand{\norm}[1]{\left\lVert#1\right\rVert}

\definecolor{cvprblue}{rgb}{0.21,0.49,0.74}
\usepackage[pagebackref,breaklinks,colorlinks,allcolors=cvprblue]{hyperref}
\def\paperID{9213} 
\def\confName{CVPR}
\def\confYear{2025}

\title{Robust SG-NeRF: Robust Scene Graph Aided Neural Surface Reconstruction}

\author{
Yi Gu \qquad Dongjun Ye \qquad Zhaorui Wang \qquad Jiaxu Wang \qquad Jiahang Cao \qquad Renjing Xu\\
HKUST(GZ)\\
{\tt\small \{ygu425, zwang408, jwang457, jcao248, renjingxu\}@connect.hkust-gz.edu.cn}\\{\tt\small imath@omnispaceai.com}\\
\url{https://rsg-nerf.github.io/RSG-NeRF/}\\
}

\begin{document}
\maketitle
\begin{abstract}
Neural surface reconstruction relies heavily on accurate camera poses as input. Despite utilizing advanced pose estimators like COLMAP or ARKit, camera poses can still be noisy. Existing pose-NeRF joint optimization methods handle poses with small noise (inliers) effectively but struggle with large noise (outliers), such as mirrored poses. In this work, we focus on mitigating the impact of outlier poses. Our method integrates an inlier-outlier confidence estimation scheme, leveraging scene graph information gathered during the data preparation phase. Unlike previous works directly using rendering metrics as the reference, we employ a detached color network that omits the viewing direction as input to minimize the impact caused by shape-radiance ambiguities. This enhanced confidence updating strategy effectively differentiates between inlier and outlier poses, allowing us to sample more rays from inlier poses to construct more reliable radiance fields. Additionally, we introduce a re-projection loss based on the current Signed Distance Function (SDF) and pose estimations, strengthening the constraints between matching image pairs. For outlier poses, we adopt a Monte Carlo re-localization method to find better solutions. We also devise a scene graph updating strategy to provide more accurate information throughout the training process. We validate our approach on the SG-NeRF and DTU datasets. Experimental results on various datasets demonstrate that our methods can consistently improve the reconstruction qualities and pose accuracies.
\end{abstract}    
\section{Introduction}
\label{sec:intro}
\begin{figure}[!t] 
\centering
\includegraphics[width=0.98\linewidth]{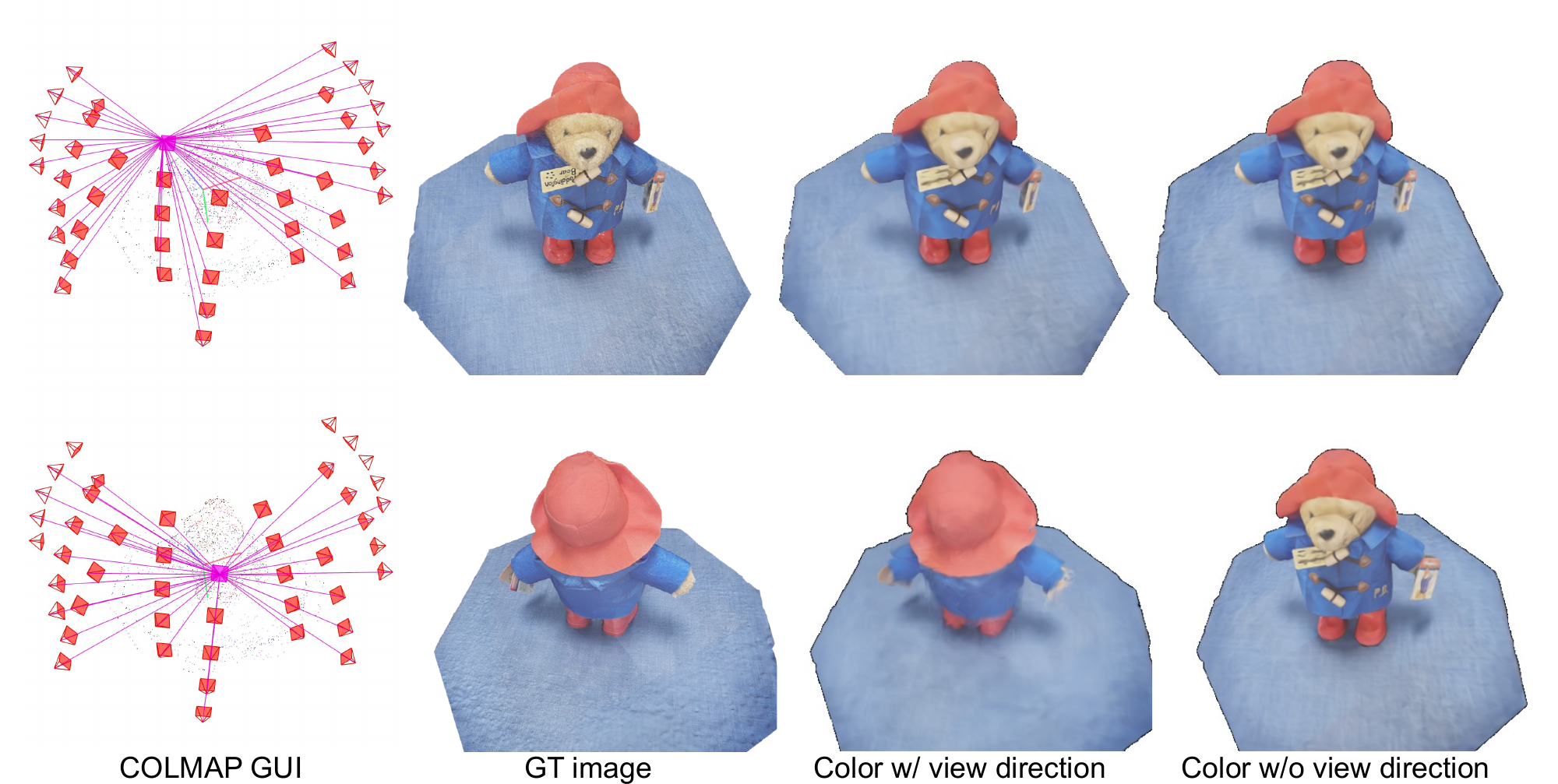}
\caption{
The illustration of the pose ambiguity. The first row is the results from inliers and the second row presents outliers. Images in the first column come from COLMAP~\cite{schoenberger2016sfm} GUI, which show that both these two poses are registered in front of the object. However, the ground truth images in the second column show the opposite phenomenon. The third column presents the rendering results of SG-NeRF~\cite{chen2024sg}, which use the same color network as NeuS~\cite{wang2021neus} with view direction as input. As shown in the fourth column, our method incorporates an isolated color network, which can well recognize this ambiguity.
}
\label{fig:shape_radiance_ambiguity} 
\end{figure}

Reconstructing the surfaces of objects from multi-view images is a fundamental challenge in both computer vision and computer graphics. Inspired by Neural Radiance Fields~\cite{mildenhall2021nerf} (NeRF), recent strides~\cite{wang2021neus, li2023neuralangelo, wu2022voxurf, miller2024objects} have marked significant progress in neural surface reconstruction (NSR) area by leveraging implicit scene representations and volume rendering techniques. In NSR, scene geometry is encoded through a signed distance function (SDF), which is learned by a multilayer perceptron (MLP) network trained with an image-based rendering loss. Despite these promising advancements, a key challenge in NSR involves the dependency on accurate camera poses. In practice, NeRF and its variants often rely on COLMAP~\cite{schoenberger2016sfm, schoenberger2016mvs}, a widely-used Structure from Motion (SfM) framework, to estimate camera poses prior. Unfortunately, these pose estimations can be significantly erroneous, adversely affecting the reconstruction quality of NeRF. Consequently, recent efforts~\cite{wang2021nerfmm, lin2021barf, jeong2021self, chng2022garf, bian2023nopenerf, chen2023local, truong2023sparf, cheng2023lu, bian2024porf} have aimed to joint optimize scene representations and camera poses to minimize the impact of pose errors. Nevertheless, most of these efforts concentrate on refining relatively small pose errors (referred to as inliers). It is still a challenge to rectify noticeably incorrect camera poses (referred to as outliers). 
\begin{figure}[!t] 
\centering
\includegraphics[width=0.98\linewidth]{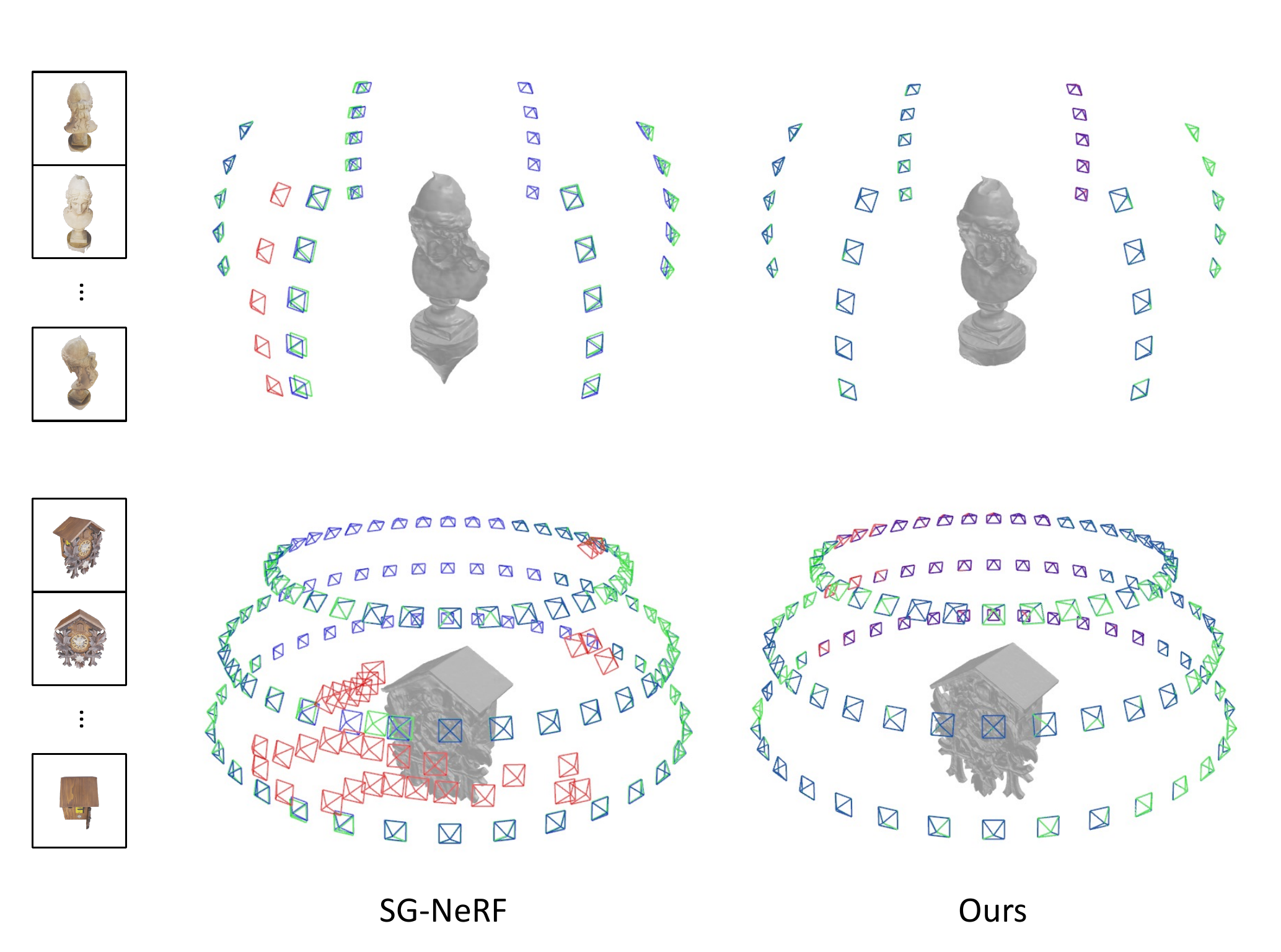}
\caption{
Reconstruction results on the SG-NeRF~\cite{chen2024sg} dataset. Both SG-NeRF~\cite{chen2024sg} and our method take the same initial poses as input, including significant noises. The camera poses are also presented with optimized \textcolor{red}{outlier poses}, \textcolor{green}{inlier poses} and \textcolor{blue}{ground truth poses}. More results are illustrated in the supplementary material.
}
\label{fig:teaser} 
\end{figure}
To alleviate the negative effects of outliers, SG-NeRF~\cite{chen2024sg} introduces scene graphs to enhance camera pose optimization for improved geometric reconstruction. The main contribution of SG-NeRF lies in estimating the confidence of each camera pose. By prioritizing ray sampling from images with high confidence poses, SG-NeRF can recover reliable geometry, even in the presence of numerous outliers, as illustrated in the left of Fig.~\ref{fig:teaser}. Theoretically, the extreme of the SG-NeRF philosophy is not sampling on outliers. Thus, it is important to recognize the inliers and outliers. However, the heuristic confidence updating strategy in SG-NeRF only depends on the peak signal-to-noise ratio (PSNR) index, which can not well reflect the differences between the inliers and outliers. As shown in Fig.~\ref{fig:shape_radiance_ambiguity}, with the wrongly estimated poses, SG-NeRF can still render images with high PSNR. This is a classical shape-radiance ambiguity problem in NeRF series~\cite{kaizhang2020nerfpp, zhu2023vdn, fang2024reducing}. A potential solution is to add some regularization terms to the loss function, but it may be more complicated coupled with a joint pose-NeRF optimization process. 

To address this problem,  we explore an improved confidence estimation method to distinguish inliers and outliers. As shown in the last column of Fig.~\ref{fig:shape_radiance_ambiguity}, we empirically find that a color network without viewing direction as input can provide more reliable information about pose confidence, albeit possibly at the expense of rendering and geometric quality. It is important to note that our primary objective is to identify inliers and outliers based on estimated confidence. To achieve this, our framework incorporates two color networks: one that aligns with traditional NSR approaches, and another is dedicated to confidence estimation. We detach the latter from the main pose-NeRF optimization graph to maintain the performance of the NSR backbone. Typically, the color network used in NSR is a shallow MLP~\cite{kaizhang2020nerfpp, mildenhall2020nerf, wang2021neus}, which means that our method does not substantially increase computational costs. This straightforward design allows us to establish a rule-based threshold to identify inliers and outliers. Subsequently, we can enhance the final results by integrating tailored designs for handling each. We present two strategies: for outliers, we employ a Monte Carlo re-localization method to provide better initialization; for inliers, we enhance constraints with re-projection and Intersection-of-Union(IoU) losses. Additionally, we devise a scene graph updating strategy based on the current SDF to eliminate incorrectly matched pairs. Experiments on the SG-NeRF~\cite{chen2024sg} and the DTU~\cite{DTU2014} datasets generally show that our method not only yields high-quality 3D reconstructions but also effectively corrects outlier poses, as illustrated in the right of Fig.~\ref{fig:teaser}. Our contributions can be summarized as follows:
\begin{itemize}
\item We propose a plug-and-play camera pose confidence estimation method that effectively identifies inliers and outliers.
\item We introduce strategies such as Monte Carlo re-localization for outliers and re-projection and IoU losses for inliers to improve geometric constraints. 
\item Additionally, we implement a scene graph updating strategy to enhance training guidance. To the best of our knowledge, this is the first study to update matching pairs dynamically during the pose-NeRF joint training process.
\end{itemize}

\section{Related works}
\label{sec:related_works}
\paragraph{Neural Surface Reconstruction.}
Traditional multi-view stereo (MVS) methods~\cite{furukawa2009accurate, galliani2016gipuma, barnes2009patchmatch, schoenberger2016mvs} explicitly establish dense correspondences across multiple images to generate depth maps, which are subsequently fused into a global dense point cloud~\cite{zach2007globally, merrell2007real}. Surface reconstruction is typically performed as a post-processing step, employing techniques such as screened Poisson surface reconstruction~\cite{kazhdan2013screened}. The processes of searching for correspondences and estimating depth have been significantly enhanced by deep learning-based approaches~\cite{yao2018mvsnet, yao2019recurrent}. Recently, the implicit representation has gained a lot of attention due to its continuity and capability to achieve high spatial resolutions. Building on the pioneering work of Neural Radiance Fields~\cite{mildenhall2020nerf} (NeRF), many successors~\cite{yariv2021volume, wang2021neus, wang2023neus2, yu2022monosdf, fu2022geo, li2023neuralangelo,wu2022voxurf} integrate the signed distance function (SDF) into NeRF to enhance geometric modeling. Among these, NeuS~\cite{wang2021neus} is particularly noteworthy for its ability to produce high-quality reconstructions and successfully handle scenes with severe occlusions and complex structures. Thus, in this study, we select NeuS to represent our scenes.

\paragraph{Structure from motion (SfM) and (re)localization.}
NeRF and its variants require accurate camera poses as input~\cite{yariv2020idr, mildenhall2020nerf, wang2021neus}. In real-world applications, Structure from Motion (SfM)~\cite{theia-manual, schoenberger2016sfm, lindenberger2021pixsfm, cui2015global, Liu_2023_LIMAP, wu2011visualsfm, snavely2008modeling, pan2024glomap} techniques are commonly employed for data pre-processing. SfM organizes a set of unstructured images by estimating camera poses and triangulating 3D scene points. An essential byproduct of this process is the scene graph, which captures information about matching pairs. However, current advanced SfM frameworks primarily depend on keypoint detection~\cite{detone2018superpoint, lowe2004distinctive, tyszkiewicz2020disk, Dusmanud2net, revaud2019r2d2} and matching~\cite{sarlin2020superglue, lindenberger2023lightglue, sun2021loftr} techniques, which can be less effective in textureless or repetitive environments. 

The task of (re)localization~\cite{sattler2016efficient, sarlin2019coarse, brachmann2021visual, dong2021robust} is also closely related to SfM. Given a database of posed images, the goal of this task is to estimate the camera poses of newly captured images. In the context of NeRF with re-localization, most existing studies~\cite{maggio2023loc, moreau2023crossfire, liu2023nerf, yen2021inerf} concentrate on relocating new images within well-constructed NeRFs. In our approach, we implement the Monte Carlo re-localization method during the training phase to improve the robustness and accuracy of outlier poses.

\paragraph{Joint NeRF and pose optimization.}
NeRFmm~\cite{wang2021nerfmm} and iNeRF~\cite{yen2021inerf} demonstrate the potential for jointly learning or refining camera parameters alongside the NeRF framework. Following works~\cite{wang2021nerf,lin2021barf,chng2022gaussian,heo2023robust,park2023camp,cheng2024improving,bian2024porf} also perform different modular modifications. For example, GARF~\cite{chng2022garf} and SiNeRF~\cite{xia2022sinerf} capitalize on the inherent smoothness of non-traditional activations to mitigate the impact of noisy gradients caused by high-frequency components in positional embeddings. L2G-NeRF~\cite{chen2023local} and Invertible Neural Warp~\cite{chng2024invertible} tackle the camera pose representation with an overparameterization strategy. NoPe-NeRF~\cite{bian2023nopenerf} employs an external monocular depth estimation model to assist in refining camera poses. Some works~\cite{truong2023sparf, bian2024porf, jeong2021self, chen2024sg} also incorporate cross-view correspondences to enhance geometry constraints. Commonly, most approaches presume that all images are properly posed initially and focus on local optimizations for pose correction. Notably similar to our method is SG-NeRF~\cite{chen2024sg}, which is the first to utilize a scene graph to guide joint optimization. Our work follows this innovative path but with a modified confidence estimation strategy.
\section{Methods}
\label{sec:methods}




\begin{figure*}[!t] 
\centering
\includegraphics[width=0.98\linewidth]{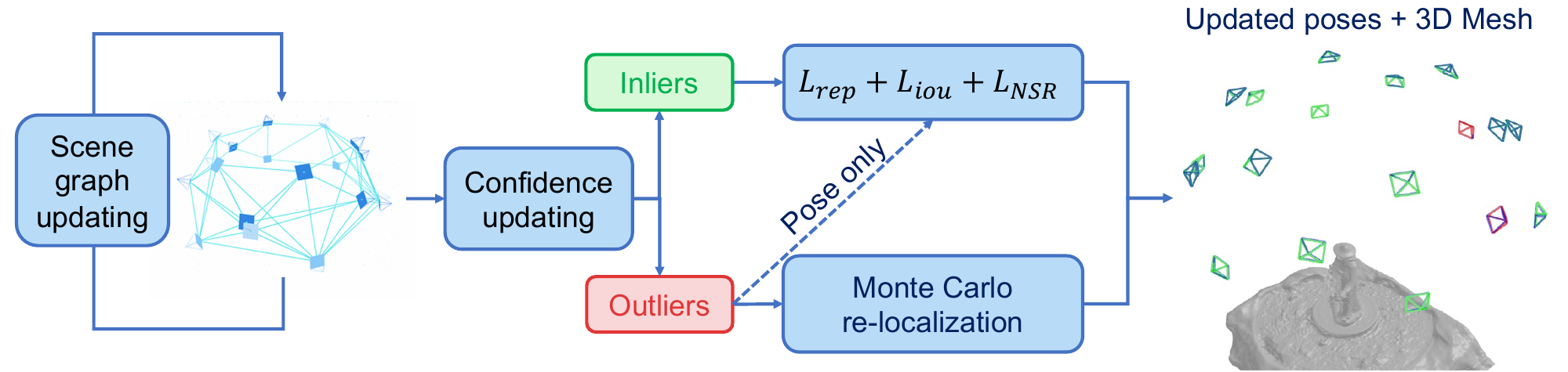}
\caption{
An overview of the proposed pipeline. Given the initial scene graph, we apply a confidence updating strategy based on an indicator from a detached color network, which can identify inlier and outlier poses. For inliers, we utilize re-projection loss and IoU loss to enhance the geometric constraints. For outliers, we utilize Monte Carlo re-localization method to find better initializations. The scene graph is also updated based on current geometry and pose estimations. Eventually, our method can reconstruct the 3D mesh from the trained field and rectify both inlier and outlier poses with high accuracy. The coloration is same as Fig.~\ref{fig:teaser}.
}
\label{fig:pipeline} 
\end{figure*}
In this section, we first define the problem setting and provide an overview of the proposed pipeline. Subsequently, we delve into the key technical designs in detail.

\paragraph{Problem statement.}
Our research focuses on the object-level 3D surface reconstruction from a set of unorganized images captured in an inward-facing configuration. Specifically, given a collection of RGB images $\mathbf{I} =\{ I_1, I_2, ..., I_N\}$, our objective is to reconstruct the 3D surface $S$ of the scene. For a specific image $I_i$, a key output of our approach is the optimized camera pose $P_i =(R_i, t_i)$, where $R_i$ belongs to $\mathbf{SO}(3)$ representing the rotation and $t_i$ is a vector in $\mathbb{R}^3$ representing the translation. Additionally, each pose is assigned an inlier-outlier confidence score.

\paragraph{Method overview.}
Fig.~\ref{fig:pipeline} illustrates the workflow of the proposed pipeline. In the data preparation stage, we first employ a widely-used Structure-from-Motion (SfM) algorithm, specifically COLMAP~\cite{schoenberger2016sfm}, to obtain the initial camera poses. Given the potential inaccuracies of these poses, proceeding with a direct joint pose-NeRF optimization could be catastrophic. To mitigate this risk, we leverage scene graph information to guide the training process (Sec.~\ref{subsec:scene_graph}). We update the confidence with our tailored indicator, which can effectively distinguish inlier and outlier poses. For inliers, we introduce additional constraints to enhance the geometric consistency (Sec.~\ref{subsec:joint_optimization}). For outliers, we utilize Monte Carlo re-localization to find better initializations (Sec.~\ref{subsec:scene_graph}). Additionally, We also devise a scene graph updating strategy to enhance the guidance during training (Sec.~\ref{subsec:scene_graph_updating}).

\subsection{Scene graph guided confidence estimation}
\label{subsec:scene_graph}

A scene graph $ \mathbf{G} = (\mathbf{V},\ \mathbf{E})$ in SfM consists of a set of nodes $\mathbf{V}$ and edges $\mathbf{E}$. Each node $v_i \in \mathbf{V}$ corresponds to an input image $I_i\in\mathbf{I}$, and an edge between two nodes contains the matching and co-visibility information about the corresponding images. We annotate all edges as $\mathbf{M} = \{ M_{i,j} | v_i, v_j \in \mathbf{V}\space,\ v_i \neq v_j\}$, where the set $M_{i,j}$ comprises all matched keypoint pairs between $I_i$ and $ I_j$. 

The original scene graph tends to be dense and contains many incorrect matches. Following SG-NeRF~\cite{chen2024sg}, we set an angular threshold $\tau$ for the estimated relative rotations and remove any edges exceeding $\tau$. Then, each node is assigned a confidence estimate based on this sparsified scene graph. 

The confidence score for a node $v_i$ is defined as the mean number of matching pairs, which can be computed as:
\begin{equation}
CS(v_i) = \frac{\sum_{M_{i,j} \in \mathbf{M}_i} |M_{i,j}|}{|\mathbf{M}_i|}  , 
\end{equation}
where $|\cdot|$ denotes the number of elements in a set, e.g., $|M_{i,j}|$ is the total number of matching pairs of $I_i$ and $I_j$ and $|\mathbf{M}_i|$ is the total number of edges of $v_i$. A higher score indicates that the image has a better matching quality and a higher likelihood of being an inlier. We normalize this confidence score via $CS(v_i) = CS(v_i) / \sum_{v \in \mathbf{V}} CS(v)$ to form a probability distribution, which guides the training to sample more rays from poses with higher confidence. All confidence computations involve a normalization step and we omit this step in the following text for brevity.

These initial scores are derived from keypoint matches, which might lack a comprehensive understanding of the information contained in images. Thus, we adaptively update the confidence scores based on the image rendering quality. Specifically, we estimate the peak signal-to-noise ratio (PSNR) for each image according to current image rendering loss for efficiency. Then, the confidence scores are updated by~\cite{chen2024sg}:

\begin{equation}
\label{eq:psnr}
CS(v_i) = CS(v_i) + \lambda_c PSNR(v_i).
\end{equation}

However,  as shown in Fig.~\ref{fig:shape_radiance_ambiguity}, we empirically found that the PSNR of outliers can be even larger than that of inliers, which means that more outliers will be sampled. The reason behind this phenomenon comes from shape-radiance ambiguity~\cite{kaizhang2020nerfpp, wang2021nerfmm, zhu2023vdn, fang2024reducing}. 

To solve this problem, we employ a new color network $C_{n}$ which does not take viewing direction as input. To mitigate the shape-radiance ambiguity and prevent overfitting, we use the same sampling points and geometry features as the original color network $C_{o}$. We detached all relevant computations from $C_{n}$ to ensure that the loss from $C_{n}$ does not impact the main networks. Since we only require an indicator that can reflect the relative rendering qualities of the training images, the PSNR estimated by $C_{n}$ can serve the same purpose as that by $C_{o}$. As highlighted in NeRF++~\cite{kaizhang2020nerfpp}, most existing works~\cite{mildenhall2020nerf, wang2021neus} use a shallow MLP for color network, which acts as an implicit regularization. Thus, $C_{n}$ will not introduce significant computational overhead. We use the PSNR estimated from $C_{n}$ (denoted as $PSNR_n$) to update the confidence score throughout the training process.

It should be noted that we also keep a record of the PSNR with $C_{o}$ (donated as $PSNR_o$), which can be helpful for filtering out outliers. When $PSNR_o$ and $PSNR_n$ show a significant discrepancy, it is an indication of anomalous data. Therefore, we recognize poses with $|PSNR_o-PSNR_n|>\tau_1$ as outliers. 

\subsection{Joint optimization}
\label{subsec:joint_optimization}
We build up our framework based on NeuS~\cite{wang2021neus}. The neural surface reconstruction loss function is defined as follows:

\begin{equation}\label{eqn:nsr}
\mathcal{L}_{NSR} = \mathcal{L}_{color}(C_o) + \mathcal{L}_{color}(C_n) + \lambda\mathcal{L}_{reg}.
\end{equation}
\noindent
The $\mathcal{L}_{color}(C_o)$ represents calculating $\mathcal{L}_{color}$ by $C_o$. The $\mathcal{L}_{color}(C_n)$ is calculated by $C_n$, with gradients only backpropagated to $C_n$.
The $\mathcal{L}_{color}$ is a photometric loss:


\begin{equation}
    \mathcal{L}_{color} = \norm{\mathbf{\hat{C}} -\mathbf{C}}_1,
\end{equation}
where $\mathbf{\hat{C}}$ is obtained by volume rendering equation~\cite{mildenhall2020nerf, kajiya1984ray} and $\mathbf{C}$ is the ground truth color.

The $\mathcal{L}_{reg}$ incorporates the Eikonal term~\cite{gropp2020implicit} applied to the sampled points to regularize the learned SDF, which can be expressed as:


\begin{equation}
\mathcal{L}_{reg} = \frac{1}{k} \sum_{i=1}^k (\norm{\nabla f(p_i)}_2 -1)^2 , 
\end{equation}
where $f(p_i)$ represents the distance estimate for each sampled 3D location along the ray.

We also utilize the Intersection-of-Union (IoU) loss $\mathcal{L}_{iou}$ and re-projection loss $\mathcal{L}_{rep}$~\cite{hartley2003multiple} to further improve the pose accuracy. The $\mathcal{L}_{iou}$ loss, firstly proposed by SG-NeRF~\cite{chen2024sg}, not only enhances geometry consistency but also accelerates convergence. Another related constraint is the epipolar loss proposed by PoRF~\cite{bian2024porf}. However, we find that both epipolar and IoU losses do not handle outliers effectively. In fact, the re-projection loss can fulfill the same role as the epipolar loss~\cite{hartley2003multiple} but is more reliable to the scene geometry. The epipolar loss does not require depth for back-projection but is invariant to the scale of the translation part. Considering the aforementioned analysis, we opt for the IoU loss and the re-projection loss in our framework.

Given a pair of matched keypoints $kp_i$ from image $I_i$ and $kp_j$ from image $I_j$, we define the Intersection Volume as:
\begin{equation}
I = MoG(kp_i) \cdot MoG(kp_j),
\end{equation}
and Union Volume as:

\begin{equation}
U = MoG(kp_i) + MoG(kp_j) - I,
\end{equation}
where $MoG(\cdot)$ is a mixture of Gaussians for sampling points along a ray. The IoU loss can be computed as:
\begin{equation}
\mathcal{L}_{iou} = 1 - \frac{I}{U}.
\end{equation}

With a set of points sampled from the ray corresponding to $kp_i$, we approximate the depth $d_i$ of $kp_i$ by selecting the point with maximal weights. Thus, the re-projection loss can be achieved by:

\begin{equation}
\mathcal{L}_{rep} = L_{\delta} \left( kp_i',\ kp_j \right) ,
\end{equation}
where $kp_i'$ is the re-projected point of $kp_i$ in image $I_j$ and $L_{\delta}$ represents the Huber loss.

We jointly optimize inlier-inlier pairs, while bypassing outlier-outlier pairs. For inlier-outlier pairs, we only optimize the poses of outliers and keep inliers and NeuS backbone fixed. Our overall loss is defined as:
\begin{equation}\label{eqn:overall}
\mathcal{L} = \mathcal{L}_{NSR} + \alpha\mathcal{L}_{iou} + \beta\mathcal{L}_{rep}.
\end{equation}


\subsection{Monte Carlo re-localization}
\label{subsec:mc_reloc}
Geometry constraints in Sec.~\ref{subsec:joint_optimization} can still struggle with certain extreme cases. One intractable case comes from the mirror-symmetry ambiguity, which has been extensively studied in SfM~\cite{ozden2010multibody, schweighofer2006robust, oberkampf1996iterative}. In the context of pose-NeRF joint optimization, NeRFmm~\cite{wang2021nerfmm} and LU-NeRF~\cite{cheng2023lu} also mentioned the same problem. LU-NeRF solves this problem by training two NeRF models, one of which uses reflected poses, requiring significant time to find the mirror poses.

Leveraging our confidence scheme, we can easily detect outliers, particularly those mirrored outliers. To maximize the use of training images, we propose to utilize Monte Carlo re-localization techniques~\cite{maggio2023loc, dellaert1999monte} to assist outliers in finding better initializations. Specifically, as we focus on inward-facing scenes, we first estimate a coarse main axis of the scene using inlier poses. The rotation around this main axis is defined as $R_{axis}(\theta)$, where $\theta$ is the angle of rotation. We then distribute the initial particles uniformly around this axis. Given a outlier pose $R_{o}, t_{o}$, the poses of these particles can be obtained by:

\begin{equation}
    R_{pi} = R_{axis}(\frac{i\cdot2\pi}{N_p}) \cdot R_{o},\ i \in \{1, 2, \dots, N_p\},
\end{equation}
and
\begin{equation}
    t_{pi} = R_{axis}(\frac{i\cdot2\pi}{N_p}) \cdot t_{o},\ i \in \{1, 2, \dots, N_p\},
\end{equation}
where $(R_{p_i}, t_{p_i})$ is the pose of $i$-th particle and $N_p$ is the number of particles. We fix all network components and only optimize the poses of particles. Initially, each particle is sampled equally for optimization. Subsequently, the sampling probability is adjusted based on the estimated $PSNR_n$. If the maximum $PSNR_n$ of the particles exceeds that of the current outlier at the end of the re-localization stage, we replace $(R_{o}, t_{o})$ with the pose of this particle. Additional details about our Monte Carlo re-localization are provided in the supplementary materials. 

\subsection{Scene graph updating}
\label{subsec:scene_graph_updating}
\begin{figure}[!t] 
\centering
\includegraphics[width=1.0\linewidth]{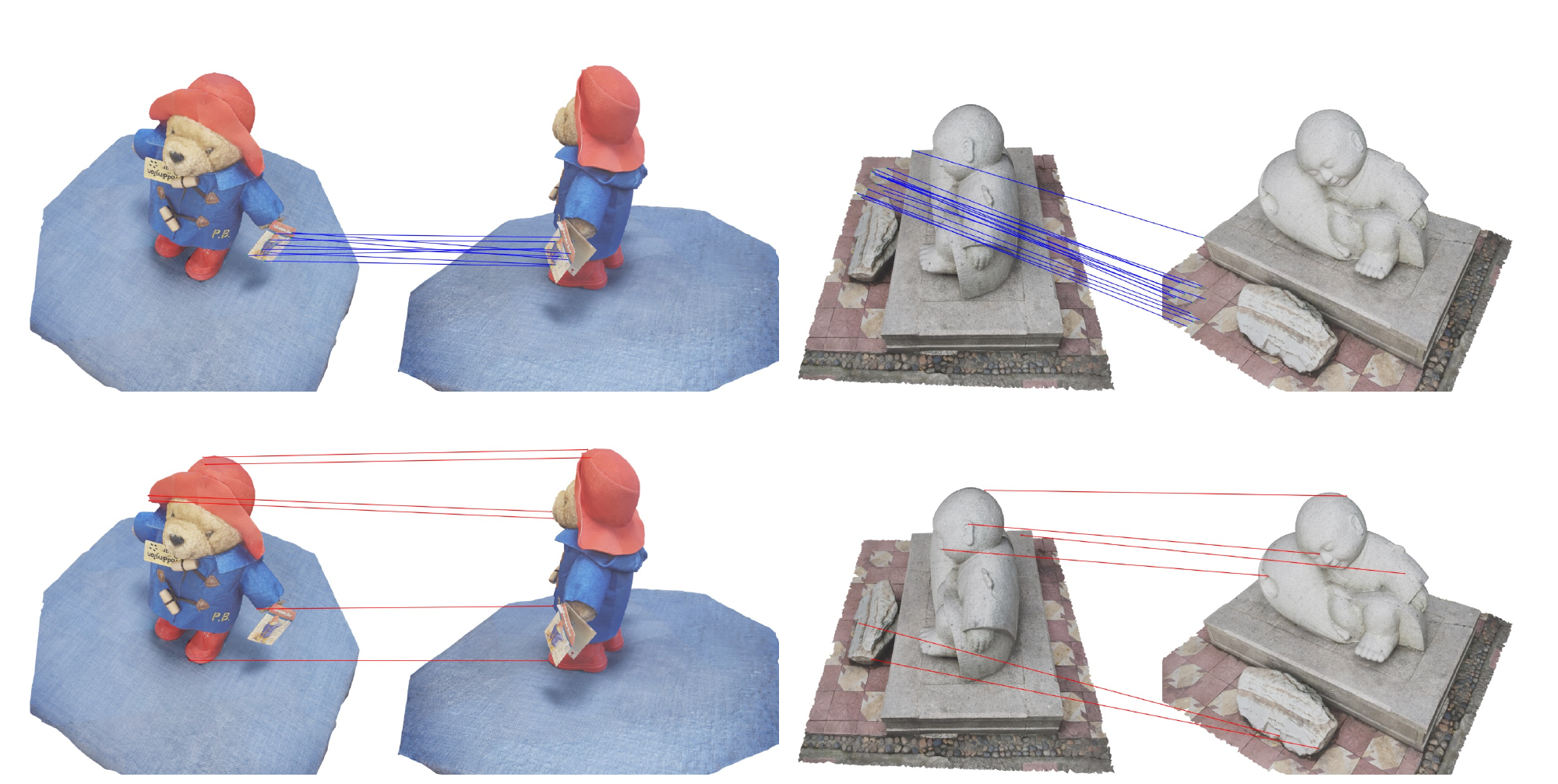}
\caption{
The illustration of scene graph updating. We filter out the wrong matched keypoint pairs (colored in red lines) and keep the correct pairs (colored in blue lines). We select image pairs with relatively few matching pairs for clearer visualization. Additional results are detailed in the supplementary materials.
}
\label{fig:updated_matching} 
\end{figure}

The initial confidence, based on results from SfM, may be sub-optimal. Therefore, we periodically update the scene graph according to current geometry and pose estimations. Similar to Sec.~\ref{subsec:scene_graph}, we use the same angular threshold $\tau$ to remove edges from the raw graph. For remaining keypoint matching pairs, we remove those with a re-projection loss surpassing the threshold $\tau_{rep}$, which is gradually reduced throughout the training. As illustrated in Fig.~\ref{fig:updated_matching}, our method effectively eliminates wrong matches, providing more reliable information for subsequent training iterations.

\section{Experiments}
\begin{figure*}[!t] 
\centering
\includegraphics[width=0.96\linewidth]{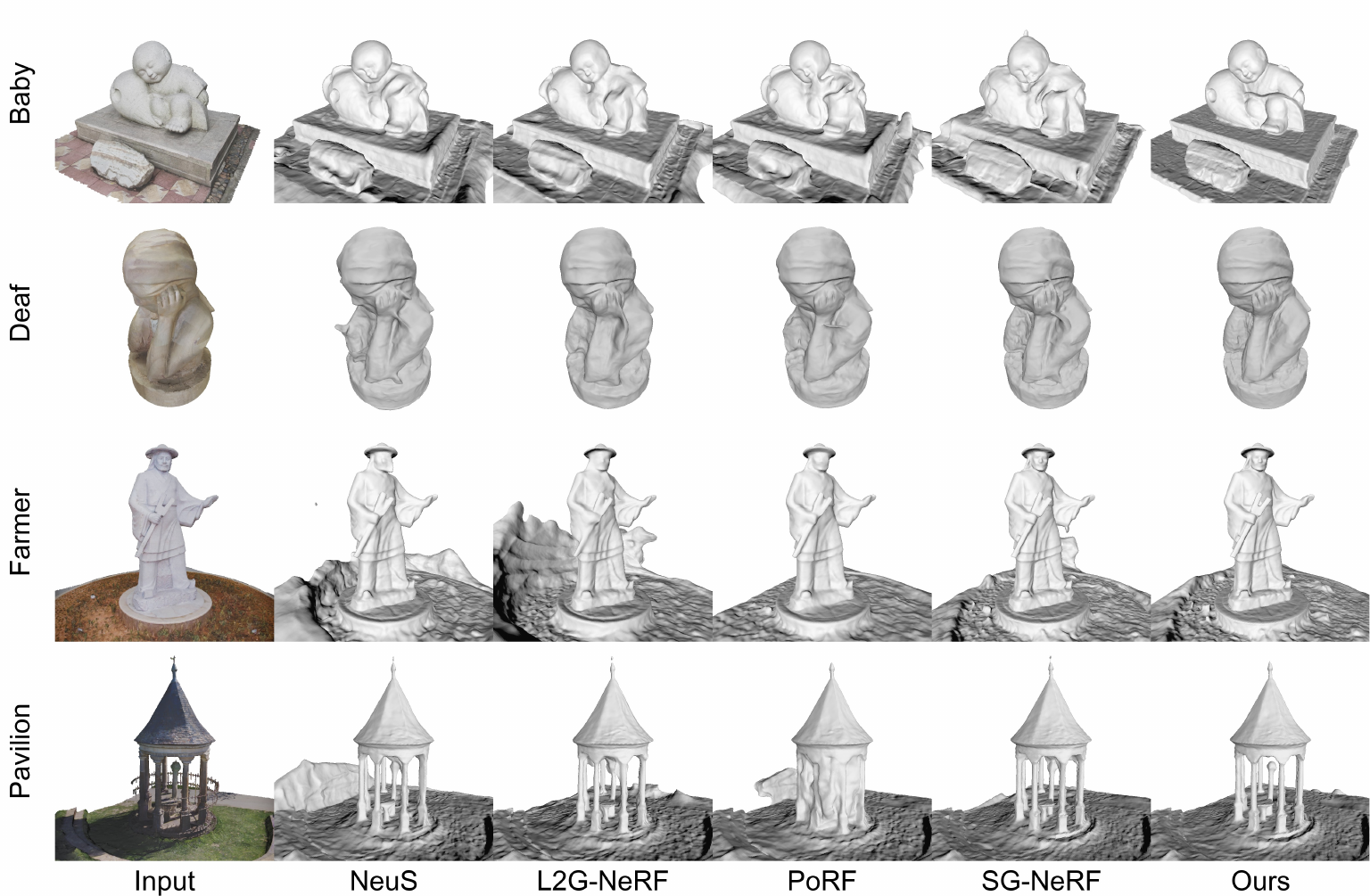}
\caption{
Qualitative comparisons on the SG-NeRF~\cite{chen2024sg} dataset. Our method can generally recover high-fidelity geometry with only one-stage training. More visual comparisons are provided in supplementary materials. 
}
\label{fig:sg_cmp} 
\end{figure*}
\subsection{Experiment setup}
Following SG-NeRF~\cite{chen2024sg}, we conduct our experiments on 8 cases from the SG-NeRF dataset and 5 cases from the DTU~\cite{DTU2014} dataset to validate our method. Following literature~\cite{wang2021neus, li2023neuralangelo}, we assess the mesh quality with Chamfer distance (CD) and F-score metrics. The baseline methods for comparison include BARF~\cite{lin2021barf}, SCNeRF~\cite{jeong2021self}, GARF~\cite{chng2022gaussian}, L2G-NeRF~\cite{chen2023local}, Joint-TensoRF~\cite{cheng2024improving}, PoRF~\cite{bian2024porf} and SG-NeRF~\cite{chen2024sg}. Results with * are achieved in a two-stage manner, including official implementations and NeuS~\cite{wang2021neus} with optimized poses. The initial camera poses of SG-NeRF are obtained by using Superpoint~\cite{detone2018superpoint} and SuperGlue~\cite{sarlin2020superglue}, with COLMAP~\cite{wu2011visualsfm, sarlin2019coarse} backend optimization. As presented in SG-NeRF~\cite{chen2024sg}, this combination consistently outperforms the standard COLMAP but still results in a proportion of significant incorrect poses, ranging from 1/9 to 1/3. The initial poses for DTU are obtained by conventional COLMAP first. To simulate outlier poses, SG-NeRF randomly selects 1/7 to 1/4 of the images for each scene and injects random noises to their poses. For a fair comparison, all methods, including ours, use the same initial poses as input.

\subsection{Implementation details}
We implement our method based on NeuS~\cite{wang2021neus}. The camera poses are implemented by Lietorch~\cite{teed2021tangent}, which can perform backpropagation on $\mathbf{SE}(3)$ Groups. Following SG-NeRF~\cite{chen2024sg}, the angular threshold $\tau$ for scene graph sparsification is set as $70$ for SG-NeRF dataset~\cite{chen2024sg} and $45$ for DTU~\cite{DTU2014} dataset, respectively. The inlier-outlier threshold is set as $\tau_1 = 9$, which is an extremely large performance gap for $PSNR_o$ and $PSNR_n$. The weights of loss are set as $\lambda=0.1, \alpha=0.2$, and $\beta=0.001$ respectively. The particle number $N_p$ is set as $24$ for efficiency. Other configurations are kept the same as NeuS. All experiments are conducted on NVIDIA RTX 3090 GPUs. Our method runs an average of 13 hours for 150k iterations on the SG-NeRF dataset, and 22 hours for 300k iterations on the DTU dataset.
\begin{table*}[!htb]
\caption{
Quantitative results on SG-NeRF~\cite{chen2024sg}. The \textbf{\textcolor{red}{red}} and \textbf{\textcolor{blue}{blue}} numbers indicate the first and second performer for each scene. $\dagger$ denotes that only valid values are used for the average. Methods with * are trained in a two-stage manner.
}

\centering
\resizebox{0.96\textwidth}{!}{ 
\begin{tabular}{cl|cccccccc|c}

\toprule
 & & Baby & Bear & Bell & Clock & Deaf & Farmer & Pavilion & Sculpture & Mean \\ 

\midrule
 
\multirow{8}{*}{\rotatebox{90}{Chamfer distance $\downarrow$}}  
& NeuS~\cite{wang2021neus} & 0.69 & 0.31 & 3.33 & 1.16 & 0.55 & 2.49 & 0.29 & 0.66 & 1.18 \\ 
& Neuralangelo~\cite{li2023neuralangelo} & 0.70 & 0.65 & - & 0.38 & 0.59 & 4.89 & 1.95 & 0.31 & 1.35$^\dagger$ \\ 
& BARF~\cite{lin2021barf}* & 1.08 & 0.28 & 3.31 & 0.19 & 0.46 & 2.13 & 0.38 & 0.57 & 1.05 \\ 
& SCNeRF~\cite{jeong2021self}* & 1.19 & 0.27 & 3.74 & 1.33 & 0.46 & 1.45 & 0.23 & 0.81 & 1.19 \\ 

& GARF~\cite{chng2022gaussian}* & 2.04 & 2.25 & 3.08 & 2.01 & 0.59 & 1.58 & 0.96 & 0.57 & 1.64 \\ 

& L2G-NeRF~\cite{chen2023local}* & 1.15 & 0.29 & 1.26 & 0.24 & 0.40 & 2.18 & - & 4.36 & 1.41$^\dagger$ \\

& Joint-TensoRF~\cite{cheng2024improving}* & 3.11 & - & 2.49 & 0.36 & 0.88 & 2.51 & 1.35 & 0.70 & 1.63$^\dagger$\\

& PoRF~\cite{bian2024porf} & \textbf{\textcolor{blue}{0.31}} & 0.49 & - & - & \textbf{\textcolor{blue}{0.30}} & 3.80 & 2.20 & - & 1.42$^\dagger$ \\

& SG-NeRF~\cite{chen2024sg} & 0.56 & \textbf{\textcolor{blue}{0.25}} & \textbf{\textcolor{red}{0.98}} & \textbf{\textcolor{blue}{0.15}} & 0.45 & \textbf{\textcolor{blue}{0.87}} & \textbf{\textcolor{blue}{0.20}} & \textbf{\textcolor{blue}{0.22}} & \textbf{\textcolor{blue}{0.46}} \\ 

& Ours & \textbf{\textcolor{red}{0.07}} & \textbf{\textcolor{red}{0.09}} & \textbf{\textcolor{blue}{1.22}} & \textbf{\textcolor{red}{0.15}} & \textbf{\textcolor{red}{0.13}} & \textbf{\textcolor{red}{0.62}} & \textbf{\textcolor{red}{0.17}} & \textbf{\textcolor{red}{0.09}} & \textbf{\textcolor{red}{0.32}} \\ 

\midrule

\multirow{8}{*}{\rotatebox{90}{F-score $\uparrow$}} 
& NeuS~\cite{wang2021neus} & 0.65 & 0.93 & 0.48 & 0.72 & 0.84 & 0.54 & 0.93 & 0.70 & 0.74 \\ 
& Neuralangelo~\cite{li2023neuralangelo} & 0.57 & 0.80 & - & 0.85 & 0.66 & 0.14 & 0.47 & 0.89 & 0.63$^\dagger$ \\ 
& BARF~\cite{lin2021barf}* & 0.58 & 0.91 & 0.49 & 0.95 & 0.86 & 0.51 & 0.86 & 0.87 & 0.75 \\ 
& SCNeRF~\cite{jeong2021self}* & 0.56 & 0.93 & 0.49 & 0.69 & 0.86 & 0.59 & \textbf{\textcolor{red}{0.95}} & 0.73 & 0.72 \\ 

& GARF~\cite{chng2022gaussian}* & 0.18 & 0.21 & 0.50 & 0.27 & 0.78 & 0.57 & 0.41 & 0.83 & 0.47 \\ 

& L2G-NeRF~\cite{chen2023local}* & 0.58 & 0.92 & 0.65 & 0.92 & 0.89 & 0.49 & - & 0.21 & 0.67$^\dagger$ \\

& Joint-TensoRF~\cite{cheng2024improving}* & 0.20 & - & 0.38 & 0.84 & 0.60 & 0.24 & 0.34 & 0.63 & 0.46$^\dagger$ \\

& PoRF~\cite{bian2024porf} & \textbf{\textcolor{blue}{0.92}} & 0.78 & - & - & \textbf{\textcolor{blue}{0.92}} & 0.39 & 0.35 & - & 0.67$^\dagger$ \\

& SG-NeRF~\cite{chen2024sg} & 0.74 & \textbf{\textcolor{blue}{0.93}} & \textbf{\textcolor{red}{0.71}} & \textbf{\textcolor{blue}{0.96}} & 0.87 & \textbf{\textcolor{blue}{0.76}} & 0.94 & \textbf{\textcolor{blue}{0.92}} & \textbf{\textcolor{blue}{0.85}} \\ 

& Ours & \textbf{\textcolor{red}{0.99}} & \textbf{\textcolor{red}{0.99}} & \textbf{\textcolor{blue}{0.65}} & \textbf{\textcolor{red}{0.96}} & \textbf{\textcolor{red}{0.99}} & \textbf{\textcolor{red}{0.79}} & \textbf{\textcolor{blue}{0.94}} & \textbf{\textcolor{red}{0.99}} & \textbf{\textcolor{red}{0.91}} \\ 

\midrule

\multirow{2}{*}{\rotatebox{90}{{\small APE} $\downarrow$}} 
& SG-NeRF~\cite{chen2024sg} & 2.16 & 1.84 & 2.15 & 1.80 & 1.17 & 0.72 & 0.6 & 1.10 & 1.44 \\ 

& Ours & 0.004 & 0.004 & 0.37 & 0.003 & 0.016 & 0.018 & 0.003 & 0.006 & \textbf{\textcolor{red}{0.053}} \\ 

\midrule

\multirow{2}{*}{\rotatebox{90}{{\small RPE} $\downarrow$}} 
& SG-NeRF~\cite{chen2024sg} & 2.26 & 1.38 & 2.29 & 0.51 & 2.10 & 1.12 & 1.04 & 2.02 & 1.59 \\ 

& Ours & 0.005 & 0.004 & 0.70 & 0.002 & 0.021 & 0.022 & 0.004 & 0.008 & \textbf{\textcolor{red}{0.096}} \\

\bottomrule

\end{tabular}

}

\label{tab:sg_results}
\end{table*}
\begin{figure}[!b] 
\centering
\includegraphics[width=0.98\linewidth]{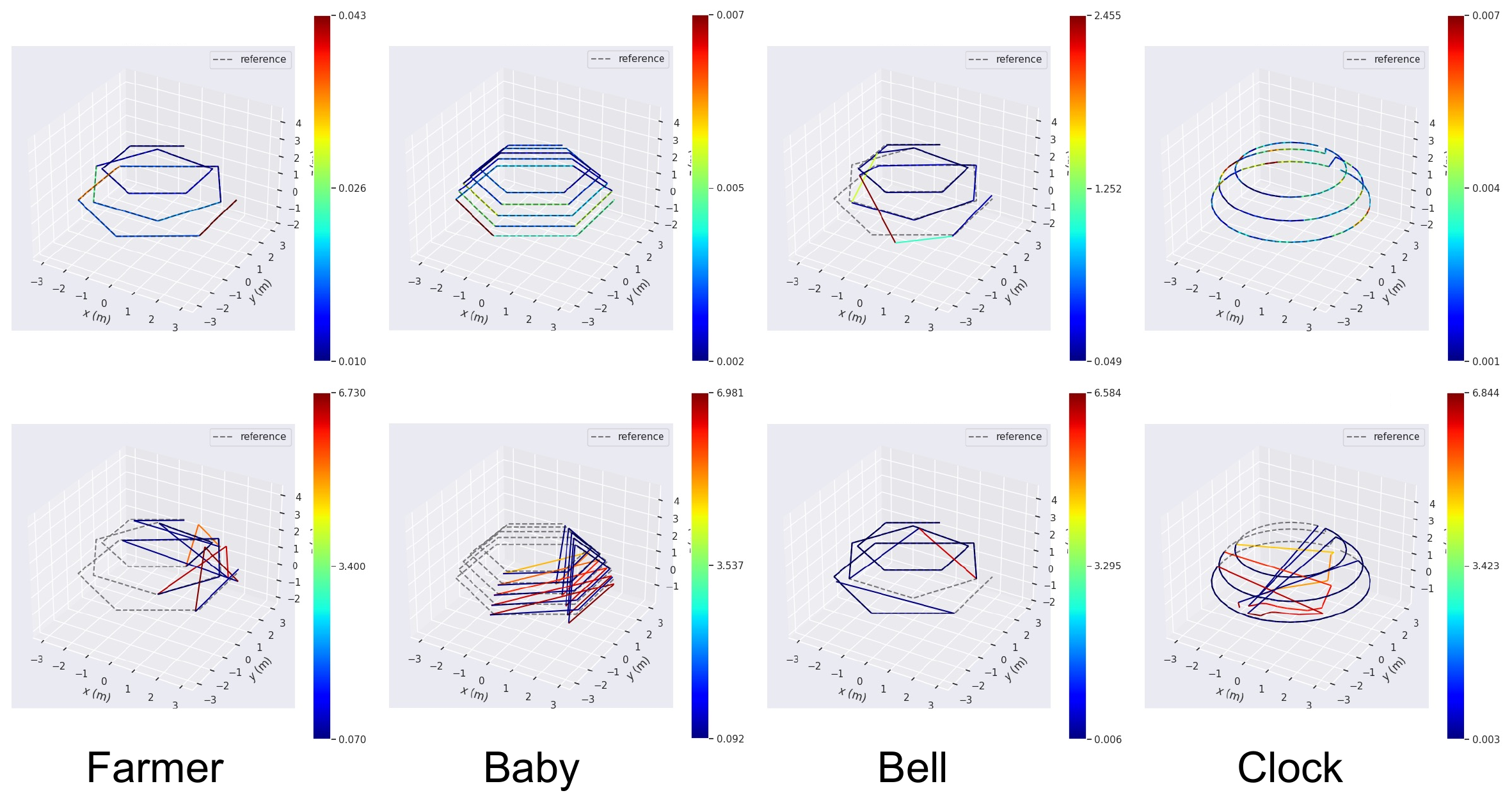}
\caption{
Visualization of pose accuracy. The fist row presents our results and the second is SG-NeRF~\cite{chen2024sg}. Dashed lines represent ground truth poses and solid lines are optimized poses. The poses rectified by our method are well aligned with the ground truth poses. In the hardest Bell case, our method can still refine most outliers, while SG-NeRF failed in all cases.
}
\label{fig:poss_acc} 
\end{figure}

\subsection{Comparisons}
\paragraph{Results on SG-NeRF.} The quantitative results are reported in Table~\ref{tab:sg_results}. Both NeuS~\cite{wang2021neus} and Neuralangelo~\cite{li2023neuralangelo} degenerate severely due to significantly noisy camera poses. PoRF~\cite{bian2024porf} and SCNeRF~\cite{jeong2021self} demonstrate commendable results in certain cases, highlighting the importance of incorporating cross-view correspondences. Among the competitors, SG-NeRF~\cite{chen2024sg} achieves the best overall performance, underscoring the effectiveness of scene graph guidance. Our method consistently outperforms other approaches by a considerable margin, which shows the effectiveness of our framework. The visual comparison is provided in Fig.~\ref{fig:sg_cmp}, where our method distinctly excels in capturing finer geometric details. However, we empirically observed that all methods, including ours, struggle with the Bell scene, likely due to the sparsity of training images. 

We also utilize evo~\cite{grupp2017evo} to evaluate the pose accuracy of our method and SG-NeRF~\cite{chen2024sg}. Due to the original SG-NeRF dataset does not provide inlier-outlier information, we utilize our indicator to filter out outliers. We align inliers to ground truth poses to get a global $\mathbf{SIM}(3)$ transformation, which is then applied to all poses. The results of absolute pose error (APE) and relative pose error (RPE) w.r.t. full transformation (including both rotation and translation parts) are reported in Table~\ref{tab:sg_results}. Our pose accuracy surpasses that of SG-NeRF by more than two orders of magnitude on both RPE and APE. Fig.~\ref{fig:poss_acc} shows the visual comparison of the camera pose accuracy.

\begin{table}[!htb]

\caption{
Quantitative results on the DTU~\cite{DTU2014} dataset with noisy camera poses as input.
}
\centering

\resizebox{0.46\textwidth}{!}{
\begin{tabular}{l|ccccc|c} 
\toprule
Chamfer distance $\downarrow$ & 24 & 37 & 40 & 55 & 63 & Mean \\ 
\midrule

NeuS~\cite{wang2021neus} 
& 1.07 & 2.80 & 1.52 & 1.30 & 3.20 & 1.98 \\

Neuralangelo~\cite{li2023neuralangelo} 
& 1.06 & 2.96 & 1.22 & \textbf{\textcolor{blue}{0.42}} & 1.23 & 1.38 \\

BARF~\cite{lin2021barf}*
& 1.46 & \textbf{\textcolor{blue}{1.40}} & 5.16 & 1.78 & 1.80 & 2.32 \\

SCNeRF~\cite{jeong2021self}* 
& 1.45 & 2.84 & 2.60 & 0.78 & 1.83 & 1.90 \\

GARF~\cite{chng2022gaussian}*
& 1.18 & 2.00 & 2.61 & 2.37 & 8.74 & 3.38 \\

L2G-NeRF~\cite{chen2023local}* & 1.08 & 1.60 & 3.27 & 1.79 & 6.97 & 2.94 \\

Joint-TensoRF~\cite{cheng2024improving}* & 1.00 & 2.60 & - & - & 7.71 & 3.77$^\dagger$ \\

PoRF~\cite{bian2024porf} & 1.15 & 2.33 & 0.97 & 0.76 & 1.30 & 1.30 \\

SG-NeRF~\cite{chen2024sg}
& \textbf{\textcolor{blue}{0.87}} & 1.83 & \textbf{\textcolor{blue}{0.88}} & \textbf{\textcolor{red}{0.38}} & \textbf{\textcolor{blue}{1.13}} & \textbf{\textcolor{blue}{1.01}} \\

Ours 
& \textbf{\textcolor{red}{0.80}} & \textbf{\textcolor{red}{1.30}} & \textbf{\textcolor{red}{0.61}} & 0.44 & \textbf{\textcolor{red}{1.09}} & \textbf{\textcolor{red}{0.85}} \\

\bottomrule
\end{tabular}
}

\label{tab:dtu_results}
\end{table}

\paragraph{Results on DTU.} 
\begin{figure*}[!t] 
\centering
\includegraphics[width=0.96\linewidth]{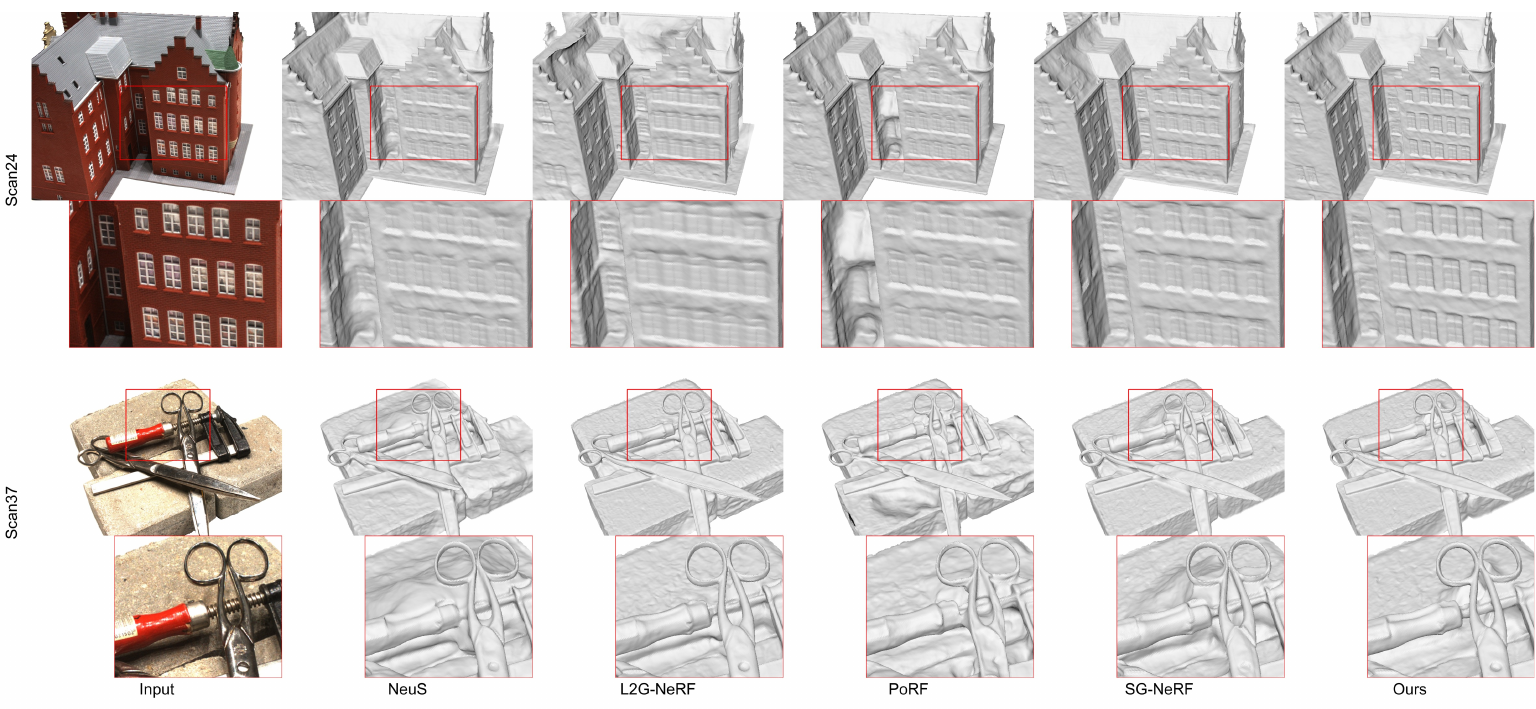}
\caption{
Qualitative comparison on the DTU~\cite{DTU2014} dataset. L2G-NeRF~\cite{chen2023local} is trained in a two-stage manner and others are trained in one stage with the same iterations. All methods take the same initial poses as input.
}
\label{fig:dtu_cmp} 
\end{figure*}
The quantitative results are shown in Table~\ref{tab:dtu_results}. We report a new result of SG-NeRF~\cite{chen2024sg} on Scan 37 with a better performance (originally reported as 2.39), due to the fact of our experiment. In DTU~\cite{chen2024sg} dataset, our method performs slightly better than SG-NeRF. We empirically find that our Monte Carlo re-localization has not been triggered. Thus, the experiment on the DTU~\cite{DTU2014} dataset can be viewed as an improved version of SG-NeRF. Our method outperforms the competitors on four scans and achieves a similar performance with SG-NeRF on Scan 55. A qualitative comparison can be found in Figure~\ref{fig:dtu_cmp}.

\begin{table}[!b]
\caption{Ablation studies on SG-NeRF~\cite{chen2024sg} and DTU~\cite{DTU2014} datasets. We individually remove scene updating (S.U.) and Monte Carlo re-localization (M.C.) on the SG-NeRF dataset. On the DTU dataset, we validate the effectiveness of the re-projection (Rep.) loss.
}

\resizebox{0.46\textwidth}{!}{
\centering
\begin{tabular}{lccc|lcc}
\toprule

\multicolumn{4}{c|}{SG-NeRF~\cite{chen2024sg}} & \multicolumn{3}{c}{DTU~\cite{DTU2014}} \\

scene & w/o S.U. & w/o M.C. & full & scene & w/o Rep. & full\\
\midrule
baby & 0.11 & 0.38 & 0.07 & scan24 & 0.91 & 0.80 \\
bear & 0.17 & 0.28 & 0.09 & scan40 & 0.78 & 0.61 \\
\bottomrule
\end{tabular}		
}

\label{tab:ablation}
\end{table}

\subsection{Ablation studies}
To validate the effectiveness of each component of our method, including the re-projection loss, Monte Carlo re-localization, and the scene updating strategy, we select 4 cases from both SG-NeRF~\cite{chen2024sg} and DTU~\cite{DTU2014} datasets. We evaluate the re-projection loss using the DTU dataset, while the other components are assessed with the SG-NeRF dataset. All experiments are conducted using our proposed confidence updating strategy. The results are reported in Table~\ref{tab:ablation}. Monte Carlo re-localization significantly enhances the results, and the scene updating strategy also contributes to a slight improvement in the final outcomes, confirming the effectiveness of these two components. Although the experiments without the re-projection loss include a cross-view constraint loss (IoU loss), our complete model still demonstrates further improvements.
\section{Conclusions}
This paper addresses neural surface reconstruction from image sets characterized by significant outlier poses. By leveraging the scene graph to guide training, we introduce a novel confidence updating strategy that effectively recognizes inliers and outliers. We enhance geometric constraints through the integration of Intersection-of-Union (IoU) loss and re-projection loss, while employing Monte Carlo re-localization techniques to accurately reposition outliers. These methods, combined with our scene graph updating strategy, enable our framework to achieve state-of-the-art performance on the challenging SG-NeRF dataset. One limitation of our approach is its dependency on a substantial number of inlier poses. As a promising direction for future research, incorporating prior models could make our framework more robust, especially in sparsely captured scenes.

{
    \small
    \bibliographystyle{ieeenat_fullname}
    \bibliography{main}
}

\clearpage
\setcounter{page}{1}
\maketitlesupplementary

\subsection{More visual results}
The interactive visual comparisons are presented in \url{https://rsg-nerf.github.io/RSG-NeRF/}.

\begin{figure}[!htbp] 
\centering
\includegraphics[width=0.98\linewidth]{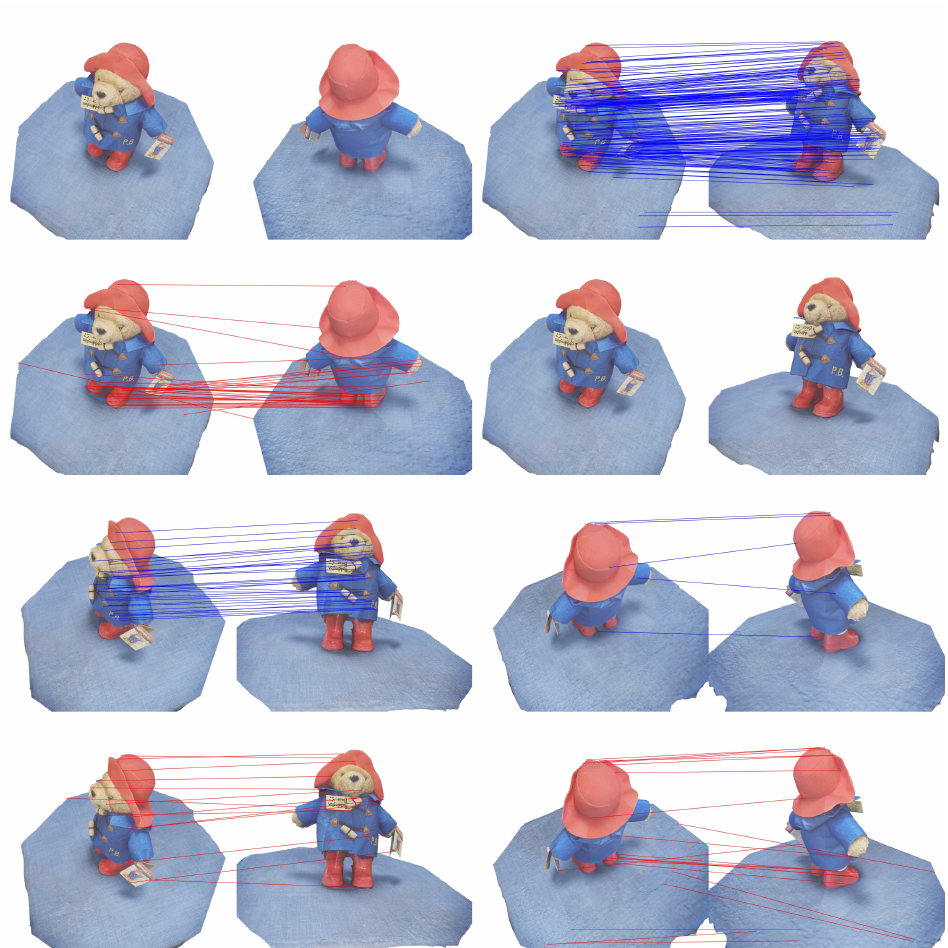}
\caption{
Scene graph updating on Bear.
}
\label{fig:sup_updated_matching} 
\end{figure}

\begin{figure}[!htbp] 
\centering
\includegraphics[width=0.98\linewidth]{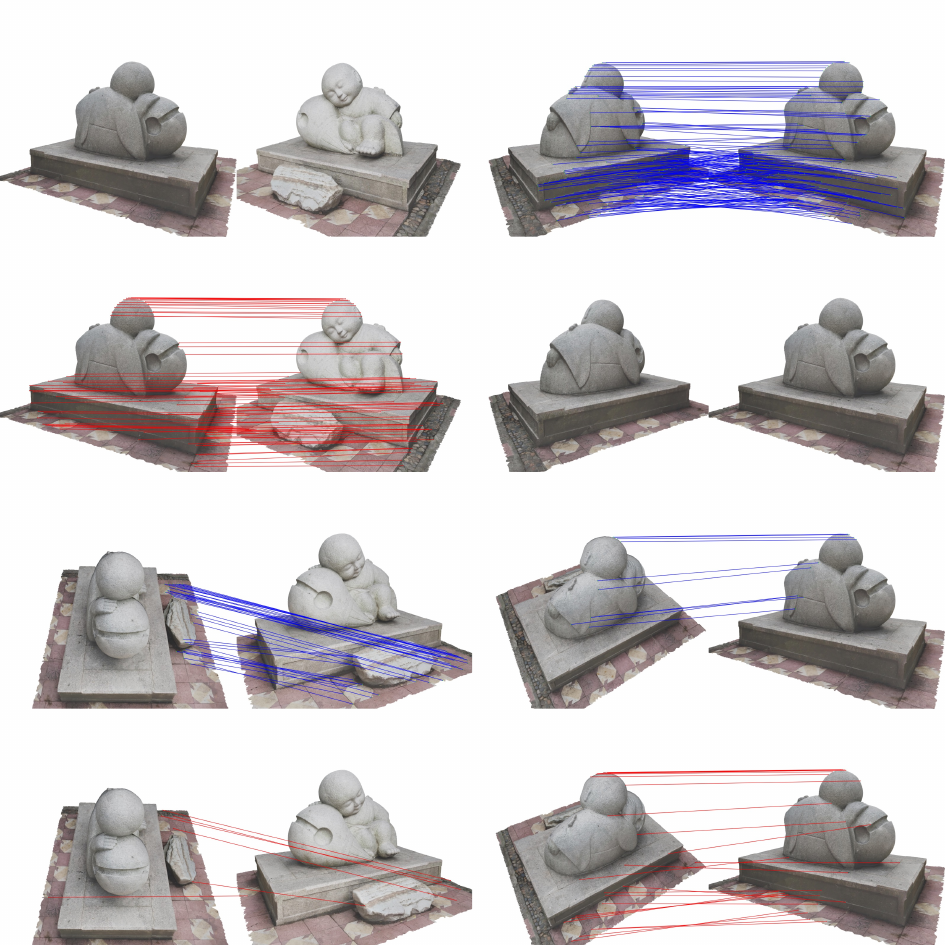}
\caption{
Scene graph updating on Baby.
}
\label{fig:sup_updated_matching_1} 
\end{figure}

Fig.~\ref{fig:sup_updated_matching} and~\ref{fig:sup_updated_matching_1} illustrate more results of the scene graph updating. We present 4 cases in Bear and Baby scenes, including all rejected, all accepted, more inlier matching and more outlier matching in each.

\begin{figure*}[!t] 
\centering
\includegraphics[width=0.98\linewidth]{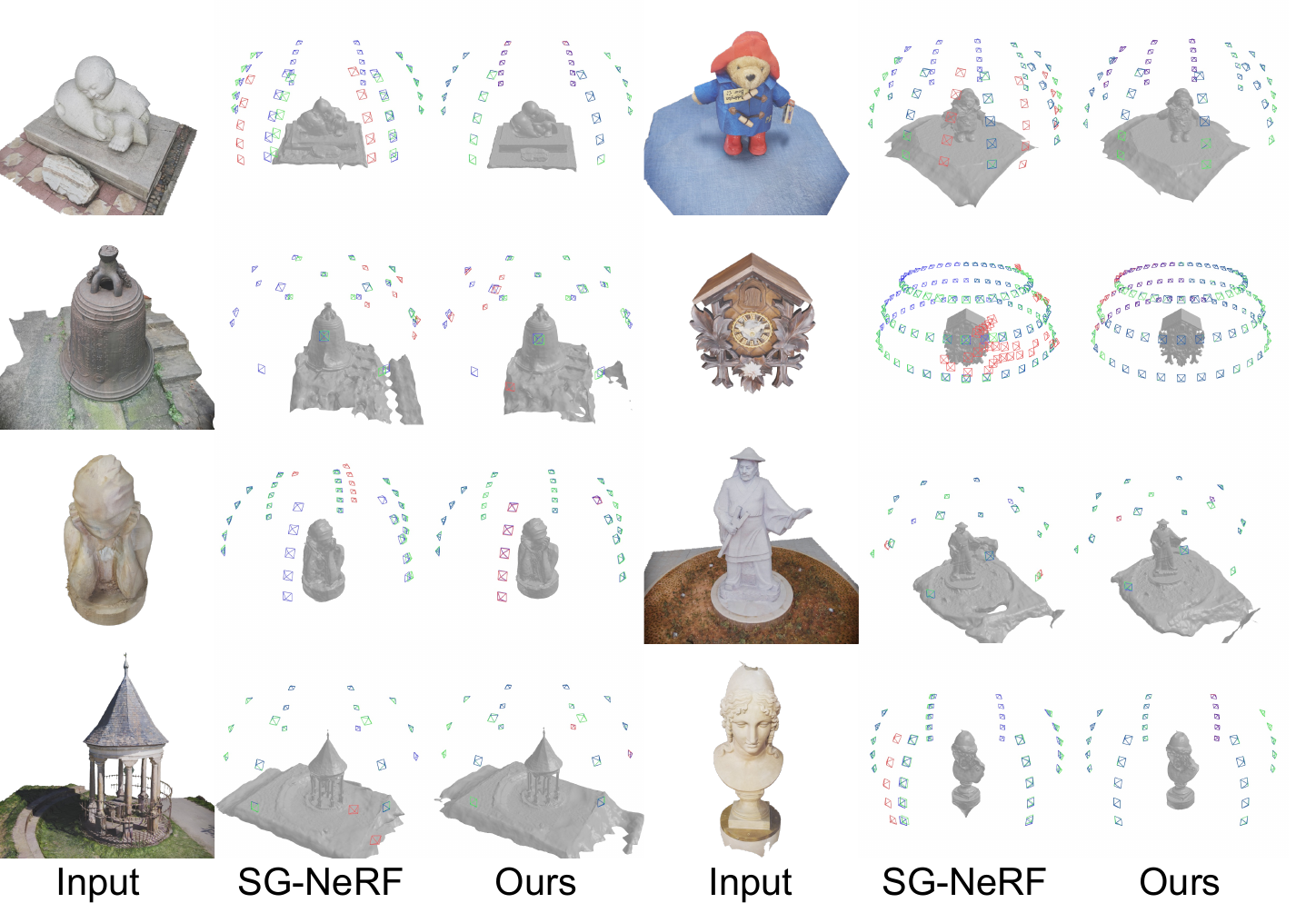}
\caption{
Reconstruction results on the SG-NeRF~\cite{chen2024sg} dataset. Both SG-NeRF~\cite{chen2024sg} and our method take the same initial poses as input, including significant noises. The camera poses are also presented with optimized \textcolor{red}{outlier poses}, \textcolor{green}{inlier poses} and \textcolor{blue}{ground truth poses}.
}
\label{fig:sup_pose_and_mesh} 
\end{figure*}

Fig.~\ref{fig:sup_pose_and_mesh} illustrate the comparisons of SG-NeRF~\cite{chen2024sg} and our method in 8 scenes of SG-NeRF dataset, including meshes and poses. 

\begin{figure*}[!t] 
\centering
\includegraphics[width=0.98\linewidth]{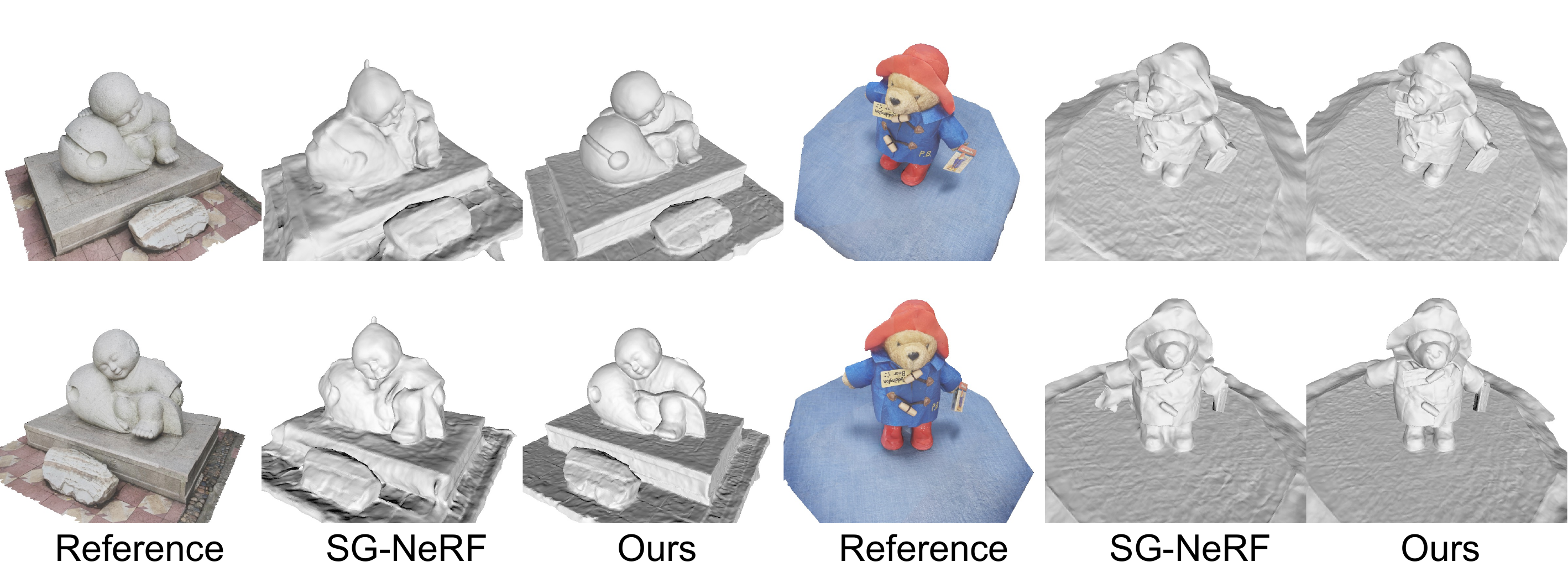}
\caption{
Reconstruction results on the SG-NeRF~\cite{chen2024sg} dataset (Baby, Bear).
}
\label{fig:sup_mesh_by2_0} 
\end{figure*}

\begin{figure*}[!t] 
\centering
\includegraphics[width=0.98\linewidth]{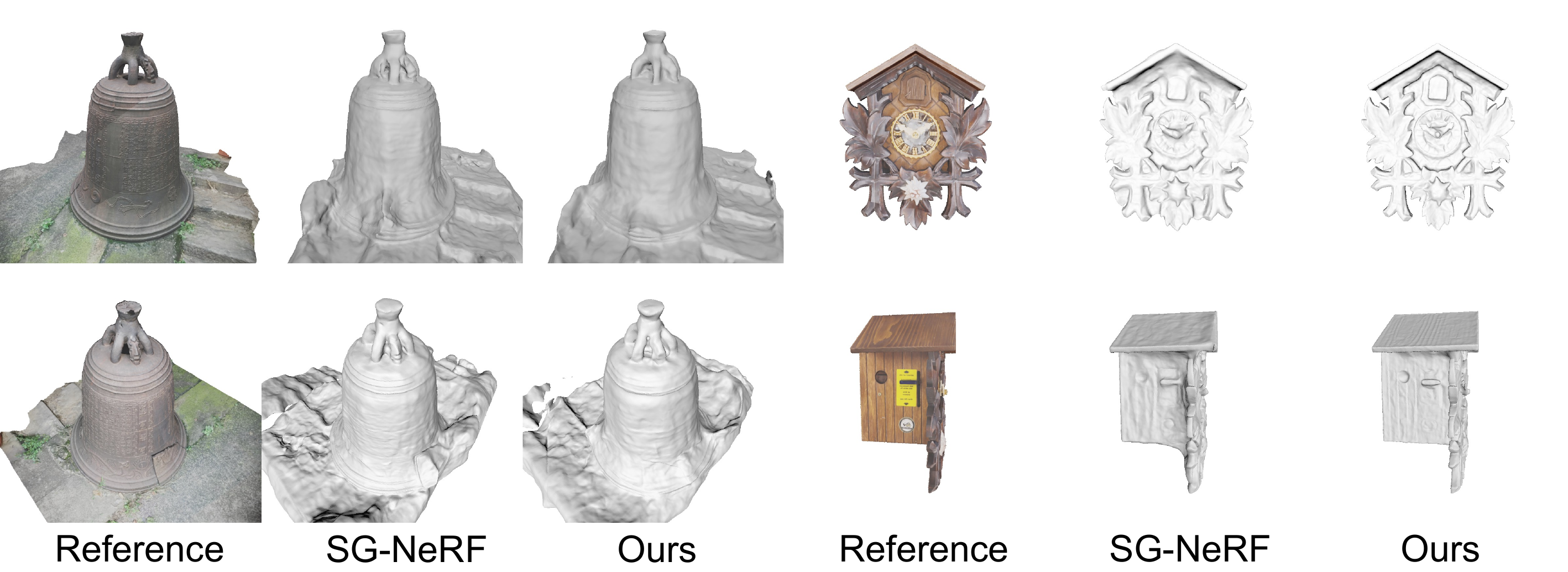}
\caption{
Reconstruction results on the SG-NeRF~\cite{chen2024sg} dataset (Bell, Clock).
}
\label{fig:sup_mesh_by2_1} 
\end{figure*}

\begin{figure*}[!t] 
\centering
\includegraphics[width=0.98\linewidth]{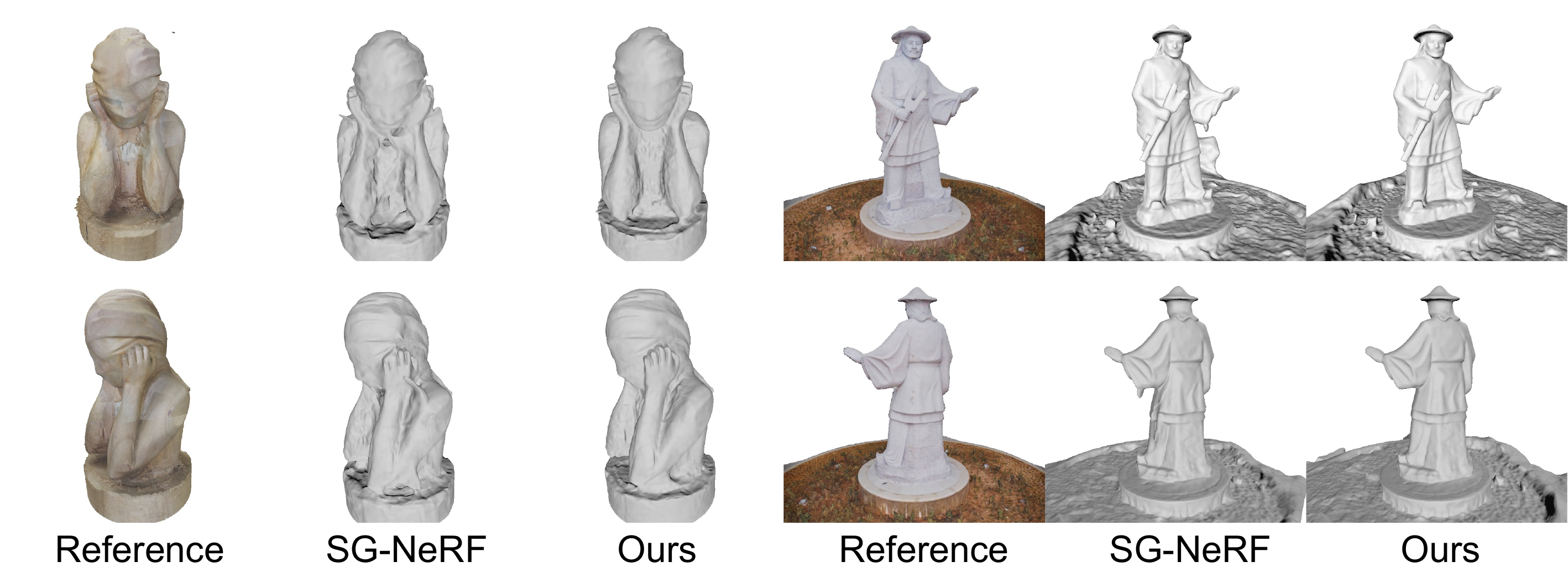}
\caption{
Reconstruction results on the SG-NeRF~\cite{chen2024sg} dataset (Deaf, Farmer).
}
\label{fig:sup_mesh_by2_2} 
\end{figure*}

\begin{figure*}[!t] 
\centering
\includegraphics[width=0.98\linewidth]{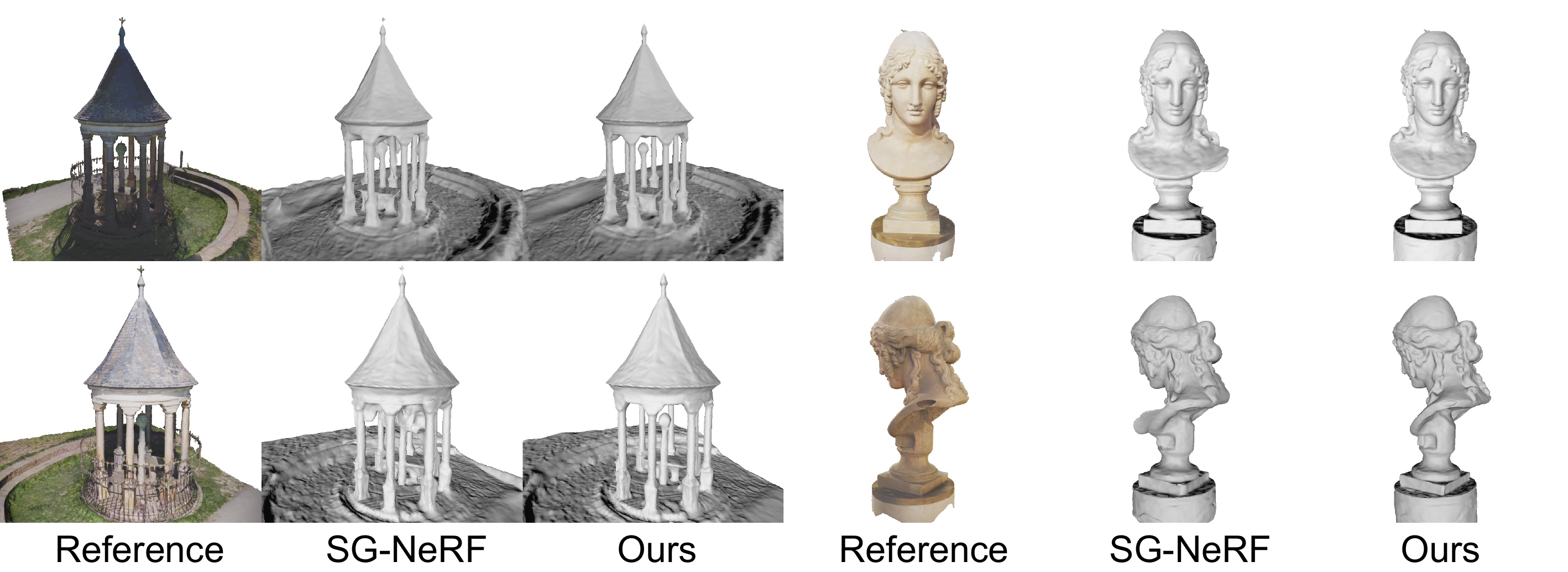}
\caption{
Reconstruction results on the SG-NeRF~\cite{chen2024sg} dataset (Pavilion, Sculpture).
}
\label{fig:sup_mesh_by2_3} 
\end{figure*}

Fig.~\ref{fig:sup_mesh_by2_0} to~\ref{fig:sup_mesh_by2_3} present the detailed mesh comparisons.

\subsection{Details about Monte Carlo re-localization}
We follow~\cite{dellaert1999monte, maggio2023loc} to implement our Monte Carlo re-localization. We fix the NeuS~\cite{wang2021neus} backbone first and utilize $\mathcal{L}_{color}(C_n)$ of optimize pose parameters. To make the gradients backpropagate to pose parameters, we remove the detach flag in the re-localization process. In the first stage, each particle is sampled with the same probability. We maintain a $PSNR_n$ list for each particle and approximate the current $PSNR_n$ by the mean value of last $10$ elements in the list. 

Subsequently, we compute the distribution of the particles according to the $PSNR_n$ in current state, abbreviated as $P_n$. For a specific particle $pi$, we define the sampling probability of $pi$ by:
\begin{equation}
    \mathcal{P}(p_i) = \frac{e^{(P_n(p_i) - \min{P_n})}}{\sum_{p\in\{p_1, p_2, ..., p_{Np}\}}{e^{(P_n(p) - \min{P_n})}}} .
\end{equation}



\end{document}